\pdfoutput=1

\documentclass[11pt]{article}

\usepackage[preprint]{acl}

\usepackage{times}
\usepackage{latexsym}
\usepackage{makecell}
\usepackage[T1]{fontenc}

\usepackage[utf8]{inputenc}

\usepackage{microtype}

\usepackage{inconsolata}

\usepackage{graphicx}
\usepackage{algorithmic}

\usepackage[ruled,vlined]{algorithm2e}
\usepackage{caption}

\usepackage{booktabs}  

\usepackage{multirow}
\usepackage{amsmath}

\usepackage{subcaption} 

\usepackage{xcolor}
\definecolor{myblue}{RGB}{70,130,180}

\title{IterSelectTune: An Iterative Training Framework for Efficient Instruction-Tuning Data Selection}

\author{
 \textbf{Jielin Song}, 
 \textbf{Siyu Liu}, 
 \textbf{Bin Zhu}, 
 \textbf{Yanghui Rao\thanks{ Corresponding author.}}
\\
School of Computer Science and Engineering, Sun Yat-sen University, Guangzhou, China
\\
\{songjlin6, liusy89, zhub35\}@mail2.sysu.edu.cn, raoyangh@mail.sysu.edu.cn
}

\begin{document}
\maketitle
\begin{abstract}


As large language models (LLMs) continue to advance, instruction tuning has become critical for improving their ability to generate accurate and contextually appropriate responses. Although numerous instruction-tuning datasets have been developed to enhance LLM performance, selecting high-quality instruction data from large source datasets typically demands significant human effort. In this work, we introduce \textbf{IterSelectTune}, an efficient, cost-effective iterative training policy for selecting high-quality instruction data with no human involvement and limited reliance on GPT-4. By fine-tuning on approximately 20\% of the source data, our method consistently outperforms models fine-tuned on the full dataset across multiple benchmarks and public test datasets. These results highlight the effectiveness of our approach in enhancing LLM performance while reducing the computational resources required for instruction tuning.

\end{abstract}

\section{Introduction}

Large Language Models (LLMs) have gained widespread recognition due to their impressive capabilities in various tasks, particularly in language generation \cite{workshop2022bloom,taylor2022galactica,touvron2023llama,zhao2023survey}. In the pretraining stage, LLMs acquire strong general abilities through next-token prediction, enabling them to excel in diverse applications. Instruction tuning \cite{LongpreHVWCTZLZ23} further enhances these models' ability to follow specific human instructions \cite{WeiBZGYLDDL22,SanhWRBSACSRDBX22,ouyang2022training,chen2023instructzero}. However, when dealing with extensive instruction datasets, fine-tuning LLMs on the whole dataset is often unnecessary, as the model may well master certain instructions. Further fine-tuning on repeated data may cause model overfitting. So the challenge lies in selecting suitable data pairs \textit{(instruction, response)} for instruction fine-tuning.


As data quality has proven to be more critical than data quantity in instruction tuning \cite{ZhouLX0SMMEYYZG23}, recent research has shifted towards selecting high-quality and diverse datasets for fine-tuning LLMs. 
While this has led to the development of methods to automate the data selection process with minimal human involvement, significant challenges remain. 
Most existing approaches rely on predefined metrics to assess data quality \cite{cao2023instruction,LiZLCC0W0024}, though effective to some extent, may not generalize well across datasets or require extensive use of GPT models like ChatGPT.

In contrast to these methods, we define high-quality instruction data as "hard" instances—those where the base LLM struggles to generate responses comparable to the original data response. Conversely, when the base LLM's response exceeds the quality of the original, it is classified as "easy" data. This approach requires a direct comparison between the base LLM's output and the original response for each instruction, offering a more tailored and direct data quality assessment that can adapt to various datasets.

However, manually performing such comparisons for large datasets is labor-intensive and requires base LLM inference for each instruction, which significantly increases time costs. While GPT-4 has been proposed as a proxy for human evaluation to reduce manual effort \cite{LiuIXWXZ23}, applying it across all data is cost-prohibitive. Therefore, our method focuses on using a smaller model in replace of GPT-4\footnote{In this study, we use the GPT-4-0125-preview version.}, minimizing its usage while maintaining high-quality data selection, making the process cost-effective and time-efficient.

In this work, we propose \textbf{IterSelectTune}, an iterative training policy framework that efficiently selects high-quality instruction data using a BERT-base \cite{DevlinCLT19} classifier. Our framework approximates GPT-4's judgment through iterative training and predicts whether a target LLM can handle an instruction effectively without needing its actual response.

The framework consists of three key components: (1) a diversity module to ensure broad coverage of instruction types, (2) an iteratively trained classifier to identify high-quality data, and (3) a similarity module that prioritizes instructions semantically close to the GPT-4-labeled "hard" data. 

The framework operates in two phases: an \textbf{iterative training phase}, where the policy is trained to replicate GPT-4’s judgments, and an \textbf{inference phase}, where the trained policy selects a portion of instruction data for fine-tuning. Our contributions are as follows:

\begin{itemize} 
    \item We introduce an iterative training policy framework that selects high-quality, diverse instruction data from large datasets with minimal GPT-4 usage and no human involvement, ensuring both cost-efficiency and scalability.
    
    \item The model fine-tuned on approximately 20\% of instruction data selected from a 120,000-instruction source dataset consistently outperforms the full-data fine-tuned model across benchmarks and test sets.

    \item In experiments with Alpaca and WizardLM, our method demonstrates strong performance with reduced data volumes (5\% of Alpaca and 10\% of WizardLM), achieving comparable results to the full-data models while requiring less time compared to other methods.
\end{itemize}

\section{Methodology}

\begin{figure*}
    \centering
    \includegraphics[width=1.0\linewidth]{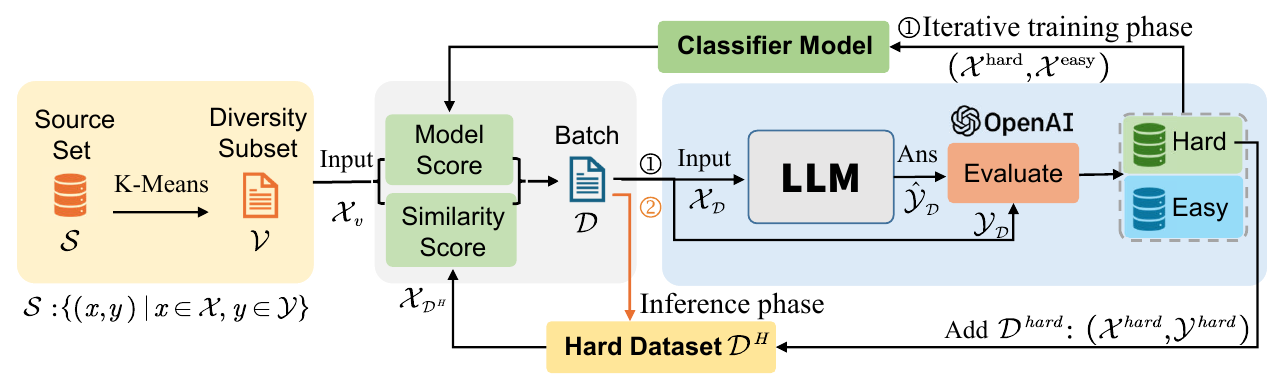}
    \caption{Illustration of our framework. We first apply K-Means clustering to the source set \(\mathcal{S}\) to derive the diversity subset \(\mathcal{V}\). Subsequently, we compute model scores and similarity scores for \(\mathcal{X}_v\), followed by sorting and selecting a batch \(\mathcal{D}\). 1) In the iterative training phase, we input \(\mathcal{X}_\mathcal{D}\) into the LLM to generate responses \(\hat{\mathcal{Y}}_\mathcal{D}\). GPT-4 then evaluates \(\hat{\mathcal{Y}}_\mathcal{D}\) and \(\mathcal{Y}_\mathcal{D}\) for binary classification. The resulting binary-classified dataset is employed to train the classifier model, enabling it to assess the quality of instructions. 2) During the inference phase, after obtaining batch \(\mathcal{D}\) through score sorting, we directly incorporate it into the hard dataset \(\mathcal{D}^\mathcal{H}\).
}
    \label{fig:model_training}
\end{figure*}

As illustrated in Figure ~\ref{fig:model_training}, our framework is divided into two main phases: iterative training phase and inference phase. Initially, we select a diverse subset of instructions from the source data. We employ a scoring mechanism that integrates classifier performance with semantic similarity to identify high-quality instructions. 
In the iterative training phase, we leverage GPT-4 to classify the instructions into "hard" and "easy" samples and use them to iteratively train the classifier. 
In the inference phase, we extract hard samples utilizing the trained classifier alongside the carefully curated "hard" samples, thereby eliminating the need for further GPT-4 involvement. 
The complete workflow is detailed in Section~\ref{subsection:The Overall Algorithm Workflow}.

\subsection{The Overall Workflow}

\label{subsection:The Overall Algorithm Workflow}

\textbf{Training Phase.}
The training process is detailed in Appendix \ref{sec:appendix_training_workflow}. 
We initiate by obtaining a diverse subset \( \mathcal{V} \) from the source set \( \mathcal{S} \) using k-means clustering. In the initial iteration, we randomly select \( \mathcal{D} \) data points without calculating scores. In subsequent iterations, we evaluate the data quality by calculating scores for the instructions in the subset and select a fixed number of high-scoring instructions \( \mathcal{D} \). These instructions are then decoded by the base LLM and subsequently evaluated by GPT-4 as either "easy" or "hard". The "hard" instructions are incorporated into the cumulative dataset \( \mathcal{D}^{\text{H}} \), while the "easy" instructions are excluded from further iterations. This labeled dataset is then employed to train the classifier, starting from the checkpoint of the previous iteration, until its validation accuracy surpasses 95\%, ensuring close alignment with GPT-4’s judgments.

To ensure cost efficiency, each iteration selects only a small batch of instructions from the large source set, minimizing the amount of GPT-4 evaluation required. 
This iterative process progressively enhances the classifier's ability to replicate GPT-4's evaluations, providing a cost-effective and labor-efficient procedure. Typically, the classifier converges after several iterations of training. Further details are provided in Appendix \ref{sec:appendix_classifier_training}.

\textbf{Inference Phase.}
The cumulative "hard" dataset $\mathcal{D}^{\text{H}}$ serves as the default high-quality subset. 
After obtaining the initial subset \( \mathcal{V} \) through k-means clustering, we proceed to score this subset using the trained classifier in conjunction with the carefully curated subset $\mathcal{D}^{\text{H}}$ for similarity. 
We then select the top \( N_{\text{sel}} \) samples based on the scores and incorporate them into $\mathcal{D}^{\text{H}}$, thereby eliminating the need for further evaluation by GPT-4. 
The algorithmic procedure is elaborated in Appendix \ref{sec:appendix_inference_workflow}.

\subsection{Diverse Subset Selection}

Ensuring data diversity is as essential as maintaining data quality in instruction tuning. A narrow focus on data from similar domains can lead to model overfitting, thereby limiting its generalization capability. 
Hence, incorporating diversity is a crucial aspect of the data selection.
In each iteration, we extract a diverse instruction subset \( \mathcal{V} \) from the source set \( \mathcal{S} \), ensuring broad representation across different sources. 
To achieve this, we apply the k-means clustering algorithm \cite{krishna1999genetic}, selecting data points from multiple clusters to promote diversity. The k-means objective function is given by:
\begin{equation}
    J = \sum_{i=1}^{k} \sum_{x \in C_i} \|x - \mu_i\|^2
\end{equation} where \( k \) denotes the number of clusters, \( C_i \) represents the data points within the \( i \)-th cluster, and \( \mu_i \) is the centroid of the \( i \)-th cluster.
Details regarding the selection of cluster numbers and data points per cluster will be discussed in Section \ref{implementation_detail}.

\subsection{Data Quality Scoring}


Following the selection of the diverse subset from the source dataset, we subsequently compute the classifier model score and the similarity score to identify high-quality instruction data that is more beneficial for fine-tuning.

\subsubsection{Classifier Model}

The classifier is a binary BERT-base model \cite{DevlinCLT19} designed to predict whether the base LLM will underperform on a given instruction. 
It classifies instructions \( x_i \) as "hard" if the base LLM's response is inferior to the original response \( y_i \), and as "easy" otherwise. 
We apply the softmax function to calculate the model score \(M(x_i)\), representing the probability that instruction \( x_i \) belongs to the "hard" category (\( y = 0 \)):
\begin{equation}
    M(x_i) = P(y=0 \mid x_i) = \frac{\exp(z_0)}{\exp(z_0) + \exp(z_1)}
\end{equation}
where the logits \( z = [z_0, z_1] \) represent the classifier's outputs for the "hard" and "easy" categories. The classifier is iteratively trained on a binary-labeled dataset updated by GPT-4 evaluations.

\subsubsection{Similarity-Based Selection}

To further enhance the selection process, we incorporate a similarity score to prioritize instructions that are semantically similar to those in the "hard" dataset \( \mathcal{D}^{\text{H}} \), thereby increasing the likelihood of selecting challenging instructions.

We utilize pre-trained BERT-based sentence encoder, bert-base-nli-mean-tokens \cite{ReimersG19}, to convert instructions into fixed-length vector representations. 
For each candidate instruction \( x_i \in \mathcal{V} \), we compute its similarity with instructions in the hard dataset \( x_h \in \mathcal{D}^H \) using cosine similarity. The similarity score \( R(x_i) \) is taken as the highest similarity value:
\begin{equation}
    R(x_i) = \max_{h \in \mathcal{D}^{\text{H}}} \text{sim}(\mathbf{v}_i, \mathbf{v}_h) = \max_{h \in \mathcal{D}^{\text{H}}} \frac{\langle\mathbf{v}_i , \mathbf{v}_h\rangle}{\|\mathbf{v}_i\| \cdot \|\mathbf{v}_h\|}
\end{equation}
where \( \mathbf{v}_i \) and \( \mathbf{v}_h \) are the vector representations of the candidate instruction and "hard" instruction, respectively.
This similarity score quantifies how closely a candidate instruction resembles a previously identified "hard" instruction, indicating its potential difficulty for the base LLM.



\subsubsection{Final Data Quality Score}

The final data quality score is a weighted sum of the classifier model score and the similarity score. This combination allows us to
account for both the likelihood that the base LLM will struggle with the instruction and its similarity to the hard dataset:
\begin{equation}
\label{eq:4}
    Q(x_i) = \alpha \cdot M(x_i) + (1 - \alpha) \cdot R(x_i)
\end{equation}
where the weighting factor \( \alpha \) balances the importance of model performance and similarity to "hard" instructions. Given that the primary objective is to prioritize model performance in determining data quality, we set \(\alpha > 0.5\). The impact of \(\alpha\) is discussed in detail in Appendix \ref{sec:alpha_study}.

\subsection{GPT-4 as a Judge}

After selecting the instruction subset \( \mathcal{D} \) based on diversity and quality, we categorize them into "easy" and "hard" labels for training the classifier.
While human evaluation is typically used for this task, it is time-consuming and costly. Instead, we leverage GPT-4 \cite{corr/abs-2303-08774}, known for its strong performance, to approximate human judgment \cite{LiuIXWXZ23,ChiangL23}.

For each instruction-response pair \( (x_i, y_i) \), where \( x_i \) is the instruction and \( y_i \) is the original response, the base model \( f_{\text{base}} \) generates a response \( \hat{y}_i \). GPT-4 compares \( \hat{y}_i \) to \( y_i \) following a predefined evaluation template (Appendix \ref{sec:prompt_evaluation}) and assigns a score \( J(\cdot) \) on a scale of 1 to 10 based on factors like accuracy and relevance.
The function \( J(\cdot) \) classifies instruction as "easy" if \( J(\hat{y}_i) > J(y_i) \), and "hard" otherwise, forming a labeled dataset:
\begin{equation}
\label{eq:5}
c_i = 
\left\{
\begin{array}{ll}
    1, & J(\hat{y}_i) > J(y_i), \\
    0, & J(\hat{y}_i) \leq J(y_i).
\end{array}
\right.
\end{equation}
where \( c_i = 1 \) indicates the instruction is easy for the base model, and \( c_i = 0 \) denotes it as hard. This labeled dataset is used to train the classifier, enabling it to approximate GPT-4's judgment in future evaluations.


To mitigate positional bias in evaluations, where the order of responses may influence scoring \cite{KoLKKK20,WangLCCZLCKLLS24}, we randomly alternate the order of responses in the training phase. Half the evaluation set is displayed in the order \( (x_i, y_i, \hat{y}_i) \), and the other half as \( (x_i, \hat{y}_i, y_i) \), reducing evaluations to one per instance and saving costs.

\section{Experimental Setup}

\subsection{Datasets}


\textbf{Training Datasets:} We compile a diverse instruction-tuning dataset by aggregating data from eight sources: Alpaca \cite{taori2023alpaca} (52,000 pairs), Dynosaur \cite{YinLYZB0C23} (802,000 pairs), Evol-Instruct \cite{LuoX0SGHT0LJ24} (70,000 pairs), LaminiLM \cite{WuWZAA24} (862,000 pairs), Dolly \cite{conover2023free} (15,000 pairs), Unnatural Instructions \cite{HonovichSLS23} (66,000 pairs), Longform \cite{koksal2023longform} (23,000 pairs), and Self-Instruct \cite{WangKMLSKH23} (82,000 pairs). We sample 15,000 instruction-response pairs from each dataset for diversity, resulting in a final source set \( \mathcal{S} \) of 120,000 examples.

\textbf{Test Datasets:} Five distinct test datasets are used for evaluation, with only their test portions employed to avoid overlap with training data. Vicuna \cite{chiang2023vicuna} (80 samples) and LIMA \cite{ZhouLX0SMMEYYZG23} (300 samples) are used for instruction following, WizardLM \cite{XuSZG0FTLJ24} (218 samples) for complex tasks, Koala \cite{geng2023koala} (180 samples) for conversational ability, and Self-Instruct \cite{WangKMLSKH23} (252 samples) for diverse instruction-following tasks.

\subsection{Implementation Details}
\label{implementation_detail}

The instruction batch size \( B \) during training is set to 400, which we consider an optimal balance between minimizing GPT-4 evaluations and ensuring effective classifier training in each iteration. The classifier is trained using an 8:2 train/valid split. 
For the diverse instruction subset \( \mathcal{V} \), we apply k-means clustering with 100 clusters, selecting 100 instruction data from each cluster to form a total of 10,000 data points per iteration. During inference, the subset size \( \mathcal{V} \) is set to three times the final selection size \( N_{\text{sel}} \), except when selecting 60\% of the source data, where \( \mathcal{V} \) is fixed at 100,000. 
This size is chosen to balance computational efficiency and data diversity. While alternative subset sizes and cluster numbers are not explored in this study, future work could examine their impact on performance. All experiments use LLaMA2-7B as the default base model. Detailed fine-tuning settings are provided in Appendix \ref{sec:appendix_implementation_details}.

\subsection{Evaluation Metrics}

\subsubsection{Evaluation on Public Test Set}

Evaluating large language models (LLMs) for instruction-following is challenging due to the diversity of valid responses and the subjectivity of human judgment. Recent advances in automated evaluation methods \cite{chang2024survey} provide scalable alternatives. In this study, we employ an LLM-based evaluation system (e.g., GPT-4) to compare outputs from two models, \( \mathcal{M}_1 \) and \( \mathcal{M}_2 \), for each instruction on the public test set. Let \( F_{\mathcal{M}_1}(z) \) and \( F_{\mathcal{M}_2}(z) \) denote the outputs of the models in response to instruction \( z \in D \), where \( D \) is the test set. A numerical score \( S(z, F_{\mathcal{M}_1}(z), F_{\mathcal{M}_2}(z)) \in [1, 10] \) is assigned based on criteria such as accuracy and relevance with template in Appendix \ref{sec:prompt_evaluation}.

To mitigate positional bias in LLM-based judgments, where the order of response presentation may affect the outcome, we apply a more comprehensive counterbalancing approach different from the training phase inspired by \cite{ChenLYWGYTS0HJ24} with two round evaluations to ensure unbiased comparisons:
In the first round, \( F_{\mathcal{M}_1}(z) \) is presented before \( F_{\mathcal{M}_2}(z) \).
In the second round, the order is reversed, with \( F_{\mathcal{M}_2}(z) \) presented before \( F_{\mathcal{M}_1}(z) \).

The model comparison adheres to the following criteria:
- \textbf{Win}: A model wins if it scores higher in both rounds or wins one round and ties the other.
- \textbf{Tie}: A tie occurs if both models receive equal scores in both rounds or one wins and one loses.
- \textbf{Loss}: A model loses if it scores lower in both rounds or ties one and loses the other.

\subsubsection{Benchmark Evaluation}

We assess the model's general reasoning and instruction-following capabilities using a range of established benchmarks from Huggingface Open LLM Leaderboard and InstructEval. For general reasoning, we evaluate with HellaSwag \cite{ZellersHBFC19}, ARC \cite{clark2018think}, TruthfulQA \cite{LinHE22}, MMLU \cite{HendrycksBBZMSS21}, RTE \cite{Poliak20}, BBH \cite{SuzgunSSGTCCLCZ23}, and DROP \cite{DuaWDSS019}. Coding ability is measured with HumanEval \cite{chen2021evaluating}.

For instruction-following tasks, we use MT-Bench \cite{ZhengC00WZL0LXZ23} for multi-turn dialogue and AlpacaEval 2.0 \cite{dubois2024length} to assess complex instruction handling.

\textbf{Settings.} We use 10-shot for HellaSwag, 25-shot for ARC, zero-shot for TruthfulQA, RTE, and HumanEval, 5-shot for MMLU, and 3-shot for BBH and DROP. MT-Bench scores are computed for both Turn 1 and Turn 2, and AlpacaEval 2.0 win rates are compared to GPT-4 Preview 1106.

\begin{figure}[ht]
    \centering
    \includegraphics[width=0.98\linewidth]{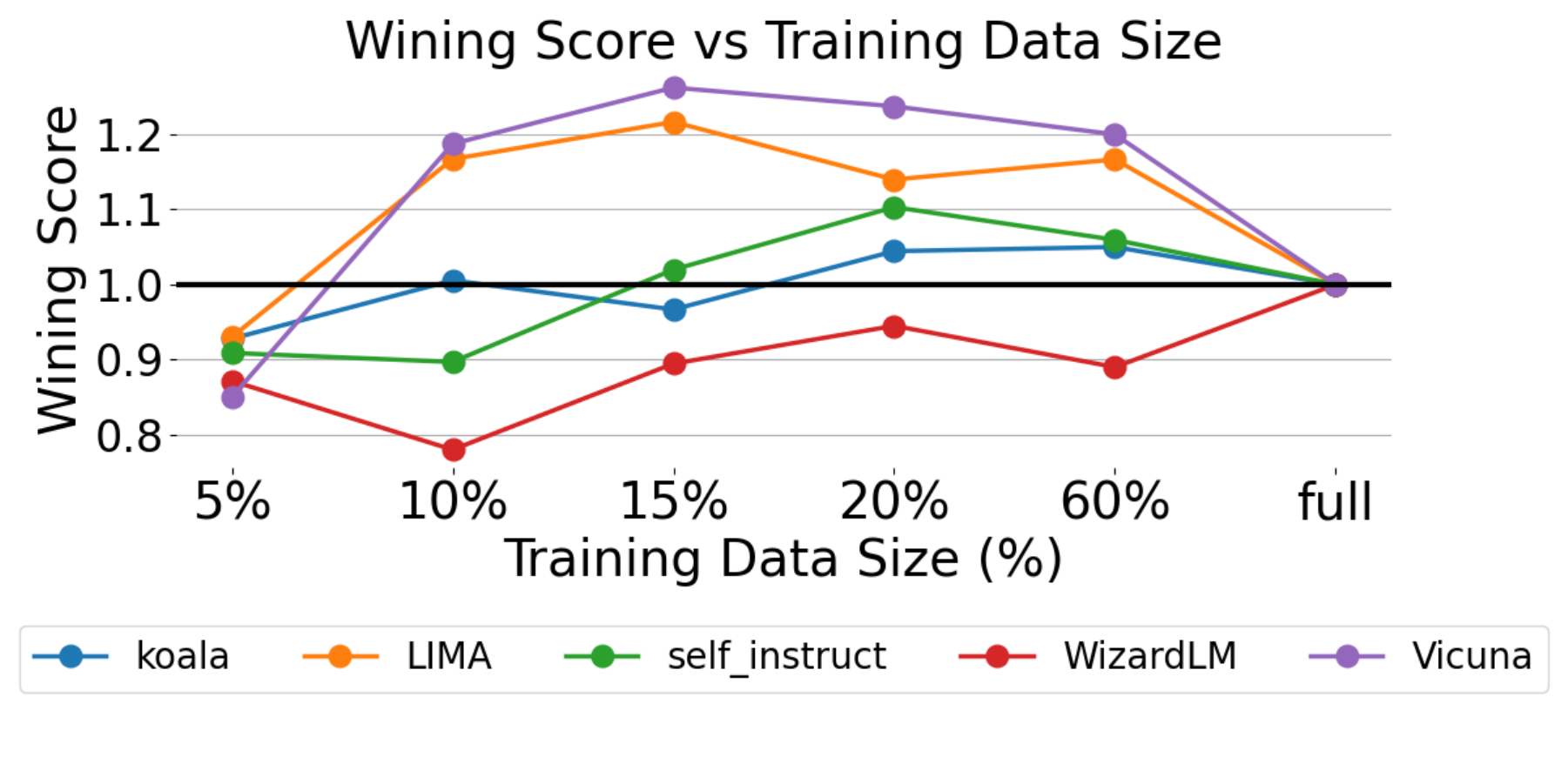} \\
    \includegraphics[width=0.98\linewidth]{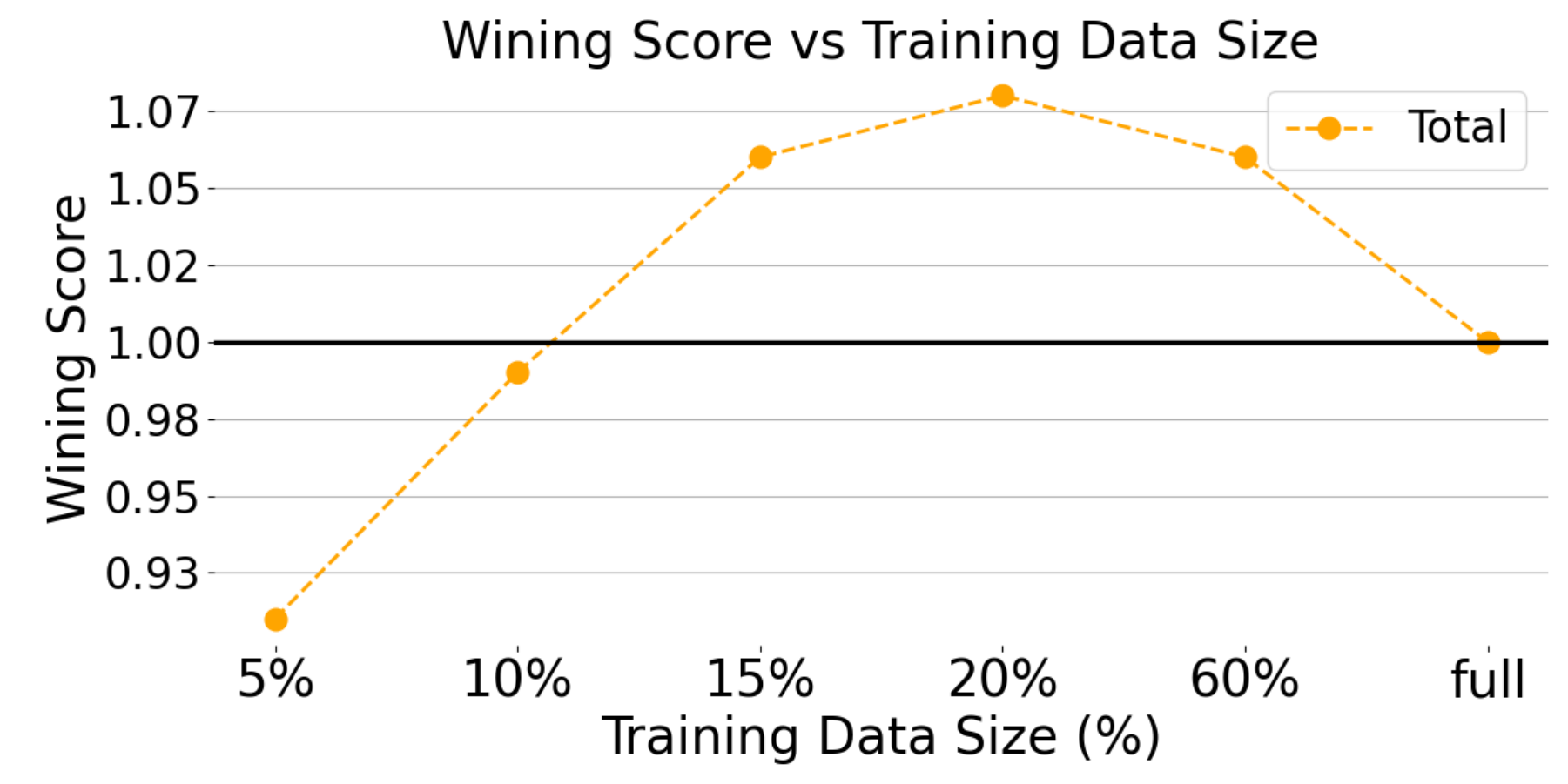}
    \caption{Winning Score vs. Training Data Size: Performance comparison across different test sets (top) and total performance (bottom).}
    \label{figure:testset_evaluation}
\end{figure}

\begin{figure}[ht]
    \centering
    \subfloat[10\% fine-tuning data]{%
        \includegraphics[width=0.98\linewidth]{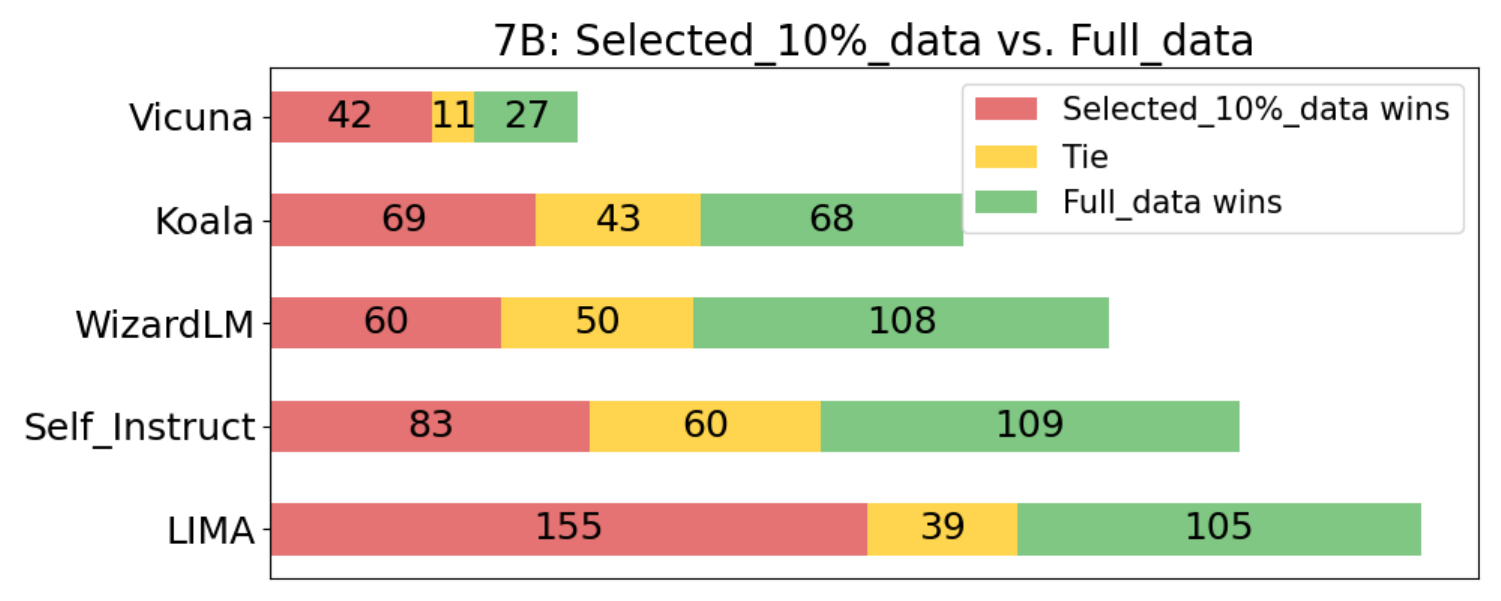}
    }\\ 
    \subfloat[20\% fine-tuning data]{%
        \includegraphics[width=0.98\linewidth]{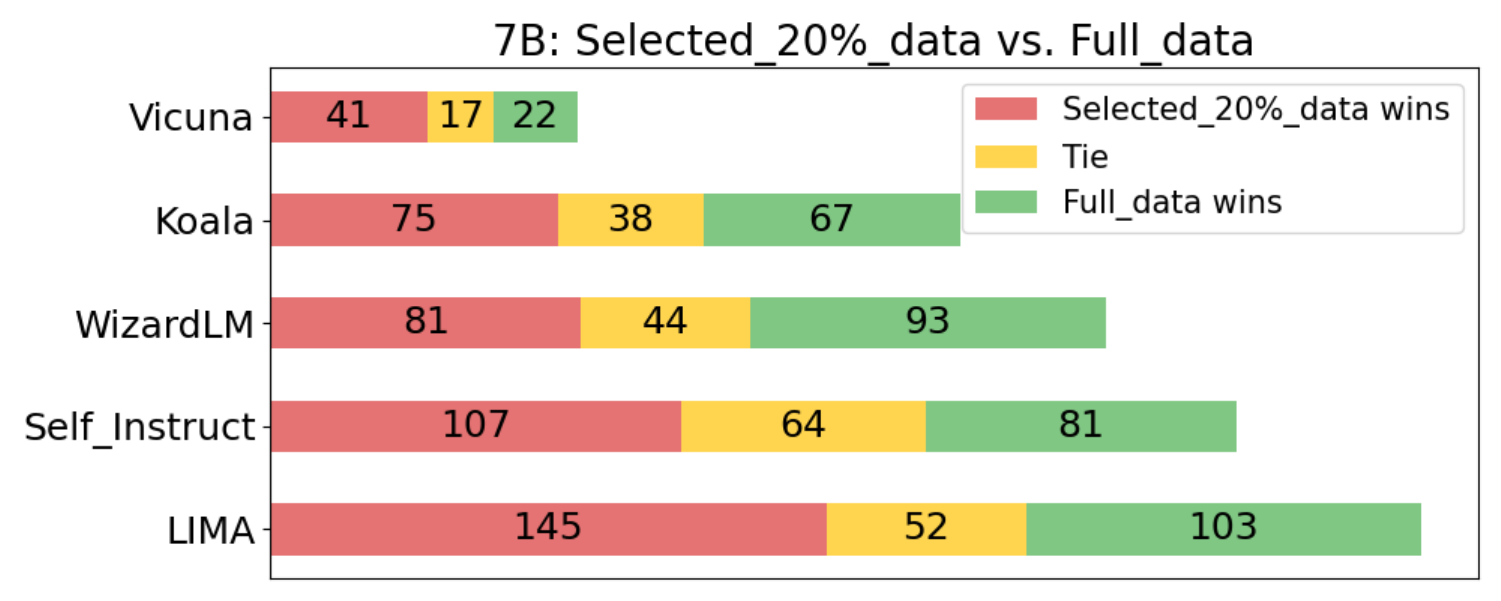}
    }
    \caption{Comparison of Win/Tie/Lose for models fine-tuned on 10\% (top) and 20\% (bottom) of the data, with the full-data fine-tuned model.}
    \label{figure:10_20_details}
\end{figure}

\section{Experimental Results}

We evaluate models fine-tuned on varying proportions of instruction-tuning data, selected through our policy using the trained classifier in inference mode from the source set \(\mathcal{S}\). We compare models fine-tuned on 5\%, 10\%, 15\%, 20\%, and 60\% of the data to a model fine-tuned on the full source set.

\subsection{Test Set Results}

Figure \ref{figure:testset_evaluation} shows model performance across individual test sets (left) and overall performance across all test sets (right). The winning score is calculated as \( \text{Winning Score} = \frac{\text{Num(Win)} - \text{Num(Lose)}}{\text{Num(TestSet)}} + 1 \), where \( \text{Num(TestSet)} = \text{Win} + \text{Tie} + \text{Lose} \). A score greater than 1 indicates that the model outperforms the full-data fine-tuned model.

As the selected data volume increases from 5\% to 20\%, performance improves across most test sets, surpassing the full-data model at 20\% on all test sets except WizardLM. However, from 20\% to 60\%, there is a performance decline, indicating that the optimal data selection portion of our policy is around 20\%. The total winning score (right plot) shows a steady improvement from 5\% to 20\%, with 15\% outperforming the full-data model and peaking at 20\%. Beyond this point, further large increases in data volume result in diminishing returns, as evidenced by the performance drop at 60\%.

Figure~\ref{figure:10_20_details} presents detailed Win/Tie/Lose comparisons for the 10\% and 20\% data scales relative to the full-data scale. The model exhibits significant improvement when increasing the data scale from 10\% to 20\% across most test sets, except for LIMA. At the 10\% data scale, the model underperforms the full-data model on most test sets. Conversely, at the 20\% data scale, it surpasses the full-data model on all test sets except WizardLM. Additional details for other data volumes are provided in Appendix \ref{sec:appendix_extra_5_15_60}.


\subsection{Benchmark Results}

We evaluate the models across several benchmarks to assess both general capabilities and instruction-following performance, comparing them to the full-data fine-tuned model.

As shown in Table~\ref{table:comparison_basic}, model performance improves as the proportion of fine-tuning data increases. From the 15\% data scale onward, the model consistently outperforms the full-data model across most benchmarks. Notably, the 20\% data fine-tuned model achieves the highest overall score, surpassing the full-data model in most tasks. However, the full-data model performs better on MMLU and BBH, likely benefiting from the larger dataset's broader knowledge and reasoning requirements.

\begin{figure}[htbp]
\centering
\includegraphics[width=0.48\textwidth]
{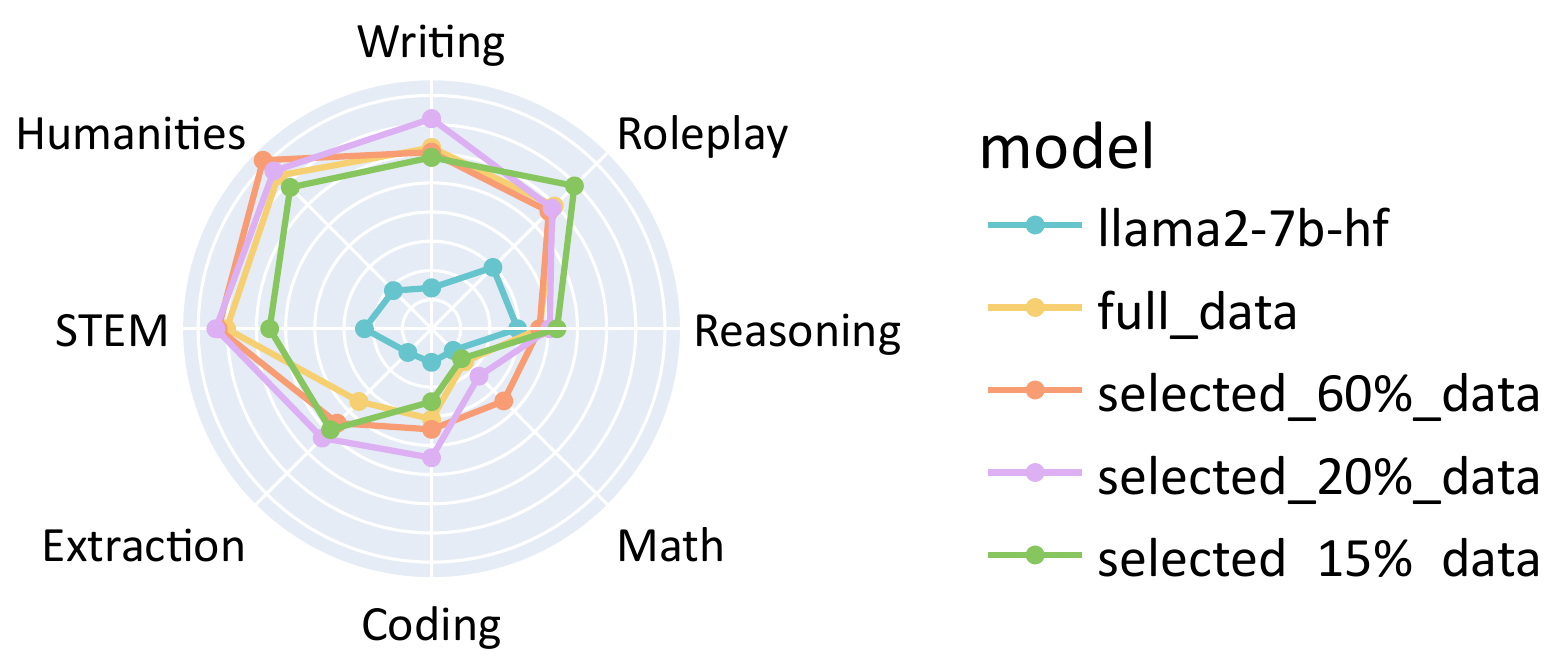}
\caption{Score visualization across multiple categories on MT-Bench.}
\label{fig:performance-chart}
\end{figure}

Table~\ref{table:comparison_instruction} presents the instruction-following benchmarks, where the 20\% data model outperforms the full-data model. Although the 60\% data model shows a slight performance drop compared to 20\%, it still exceeds the full-data model. Figure~\ref{fig:performance-chart} further illustrates that the 20\% data model achieves the best results across MT Bench categories, outperforming the full-data model on most tasks.

Across all experiments, models fine-tuned on selected data, particularly the 20\% subset, consistently outperform the full-data model, highlighting the effectiveness of our data selection framework.

The first row in each table shows the performance of the base model (LLaMA2-7B) without fine-tuning. All fine-tuned models significantly outperform the base model across every benchmark, demonstrating the positive impact of fine-tuning on model performance.

\begin{table*}[ht]
\centering
\resizebox{\textwidth}{!}{
\begin{tabular}{l|c|ccccc|ccc}
\hline
 & \multicolumn{1}{c|}{Overall} &\multicolumn{5}{c|}{Huggingface Open LLM Leaderboard} & \multicolumn{3}{c}{InstructEval} \\
 & Average & HellaSwag & ARC & TruthfulQA & MMLU & RTE & BBH & DROP & HumanEval \\
\hline
LLaMA2-7b-hf & 34.88 & 73.01 & 44.2 & 28.21 & 32.94 & 60.29 & 28.88 & 9.1 & 2.44 \\
Selected\_5\%\_data & 42.65 & 78.99 & 46.16 & 36.42 & 40.61 & 71.84 & 32.13 & 22.82 & 12.2 \\
Selected\_10\%\_data & 43.78 & 79.42 & 47.7 & 35.71 & 41.66 & 72.56 & 32.93 & 23.79 & \textbf{16.46} \\
Selected\_15\%\_data & 44.52 & 79.52 & 46.76 & 38.29 & 44.44 & 75.09 & 33.85 & 24.82 & 13.41 \\
Selected\_20\%\_data & \textbf{46.15} & \textbf{79.9} & 47.44 & \textbf{38.58} & 45.53 & \textbf{78.7} & 33.78 & 28.81 & \textbf{16.46} \\
Selected\_60\%\_data & 45.29 & 79.24 & \textbf{48.89} & 36.01 & 46.37 & 72.92 & 33.91 & \textbf{29.72} & 15.24 \\
Full\_data & 44.06 & 79.17 & 48.72 & 34.34 & \textbf{46.45} & 71.12 & \textbf{34.07} & 25.84 & 12.8 \\
\hline
\end{tabular}
}
\caption{The model performance on Huggingface Open LLM Leaderboard and InstructEval Leaderboard.}
\label{table:comparison_basic}
\end{table*}

\begin{table}[ht]
\centering
\resizebox{0.5\textwidth}{!}{ 
    \begin{tabular}{l|ccc|cc}
    \hline
     & \multicolumn{3}{c|}{MT Bench} & \multicolumn{2}{c}{AlpacaEval 2.0} \\
     & Overall & turn1 & turn2 & \makecell{length controlled \\ win rate} & win rate \\
    \hline
    LLaMA2-7b-hf & 1.814 & 2.084 & 1.521 & - & - \\
    Selected\_10\%\_data & 4.596 & 5.456 & 3.736 & 3.9 & 1.91 \\
    Selected\_15\%\_data & 4.756 & 5.881 & 3.631 & 3.69 & 1.95 \\
    Selected\_20\%\_data & \textbf{5.228} & \textbf{6.194} & \textbf{4.263} & \textbf{4.92} & \textbf{2.65} \\
    Selected\_60\%\_data & 4.941 & 5.956 & 3.925 & 3.6 & 2.13 \\
    Full\_data & 4.817 & 5.434 & 4.2 & 4.03 & 2.01 \\
    \hline
    \end{tabular}
}
\caption{The model performance on MT Bench and AlpacaEval 2.0.}
\label{table:comparison_instruction}
\end{table}

\section{Results on Alpaca and WizardLM Models}

To further validate our method, we conduct experiments with Alpaca \cite{taori2023alpaca} and WizardLM \cite{XuSZG0FTLJ24}, both fine-tuned on LLaMA 7B, following the experimental setup and evaluation metrics in \cite{LiZLCC0W0024}.


Although the base LLM differs from the main experiments (LLaMA2-7B), we assume that "hard" instructions for LLaMA2 would similarly challenge LLaMA, as LLaMA2 is a more advanced version. Thus, we directly apply the inference mode of our policy (implementation details in Appendix \ref{sec:appendix_alpaca_wizardlm_implementation_details}). Table~\ref{table:comparison_new_alpacaxWizardLM} compares our models' performance with the official Alpaca and WizardLM models, as well as the Instruction-Following Difficulty (IFD) results from \cite{LiZLCC0W0024}.

For the Alpaca model, fine-tuning on 5\% of the instruction data, our method outperforms \cite{LiZLCC0W0024} on most benchmarks, except for ARC and AlpacaEval 1.0, where the lag in ARC explains the minor difference in the overall average. However, we achieve notable gains on MMLU and TruthfulQA, demonstrating our method’s strength in general knowledge and factual accuracy tasks. For WizardLM, using 10\% of the instruction data, our model achieves comparable performance to reimplemented WizardLM on most benchmarks and slightly surpasses \cite{LiZLCC0W0024} in ARC and HellaSwag.

In terms of time complexity, our method requires \( \mathcal{O}(n \times \mathcal{D}) \) inferences on the base LLM, where \( \mathcal{D} \) is the number of instructions in the small batch and \( n \) is the number of training iterations. Since \( N \) represents the total number of instructions in the dataset, and the small batch size is significantly smaller than the full dataset (\( \mathcal{D} \ll N \)), with only a few iterations required (\( n \)), it follows that \( n \times \mathcal{D} \ll N \). Additionally, \( N - n\mathcal{D} \) inferences are performed using a smaller, more efficient BERT-like model, which is computationally inexpensive. Therefore, our approach significantly reduces computational cost compared to \cite{LiZLCC0W0024}, which requires \( \mathcal{O}(N) \) inferences on the base LLM.

\begin{table*}[ht]
\centering
\resizebox{\textwidth}{!}{
\begin{tabular}{l|ccccc|c|c}
\hline
 & \multicolumn{5}{c|}{Huggingface Open LLM Leaderboard} & \multicolumn{1}{c|}{AlpacaEval 1.0} & \multirow{2}{*}{Time Complexity} \\
 & Average & ARC & HellaSwag & MMLU & TruthfulQA & AlpacaEval 1.0 &  \\
\hline
Official Alpaca* & 50.21 & 42.65 & 76.91 & 41.73 & 39.55 & 26.46 & - \\
IFD (5\% Alpaca)* \cite{LiZLCC0W0024} & 52.06 & 53.92 & 79.49 & 36.51 & 38.33 & 34.74 & \( \mathcal{O}(N) \) \\
Ours (5\% Alpaca) & 51.82 & 47.53 & 79.62 & 39.69 & 40.42 & 33.85 & \( \mathcal{O}(n \times \mathcal{D}) \) \\
\hline
Reimplemented WizardLM* & 52.79 & 53.07 & 77.44 & 37.75 & 42.90 & 61.99 & - \\
IFD (10\% WizardLM)* \cite{LiZLCC0W0024} & 51.59 & 52.90 & 78.95 & 33.08 & 41.41 & 61.44 & \( \mathcal{O}(N) \) \\
Ours (10\% WizardLM) & 52.24 & 55.92 & 79.03 & 32.96 & 41.06 & 60.94 & \( \mathcal{O}(n \times \mathcal{D}) \) \\
\hline
\end{tabular}
}
\caption{Performance comparison of Alpaca and WizardLM on the Huggingface Open LLM Leaderboard and AlpacaEval 1.0. Results marked with * are taken from \cite{LiZLCC0W0024}.}
\label{table:comparison_new_alpacaxWizardLM}
\end{table*}

\section{Ablation study}

\subsection{Component Exclusion Analysis}

We conduct an ablation study to evaluate the impact of each component, with data selection fixed at 20\%. The variations tested include:

1. \texttt{diversity\_only}: Selects data using only k-means clustering to test the effect of diversity without scoring.
2. \texttt{non\_iterative}: Trains the classifier without iterative updates to evaluate the role of iterative training.
3. \texttt{random\_selection}: Randomly selects data to assess performance without guided selection.
4. \texttt{score\_only}: Selects data based solely on classifier and similarity scores, omitting diversity considerations.

Results on benchmark tasks highlight the impact of each component. In general capability benchmarks (Table~\ref{table:comparison_basic_ablation}), our method consistently outperforms others, achieving the highest scores on most tasks. \texttt{random\_selection} model performs best on ARC, likely due to ARC’s focus on factual recall, where random sampling may have favored data points better suited for this task. On TruthfulQA and RTE, both our method and \texttt{score\_only} model show significant improvement, validating the scoring mechanism. However, \texttt{score\_only} model performs noticeably worse on MMLU, demonstrating the importance of diverse data during fine-tuning. Furthermore, \texttt{non\_iterative} shows a substantial drop in DROP, highlighting the need for iterative training to refine proper data selection.

In instruction-following benchmarks (Table~\ref{table:comparison_instrucion_ablation}), our method achieves top scores on MT Bench and AlpacaEval 2.0. Both our method and \texttt{score\_only} model excel on AlpacaEval 2.0, further supporting the effectiveness of the scoring mechanism in selecting high-quality instruction data. Detailed results on test sets are provided in Appendix~\ref{sec:appendix_testset_ablation}.






\begin{table*}[ht]
\centering
\resizebox{\textwidth}{!}{
\begin{tabular}{l|c|ccccc|ccc}
\hline
 & \multicolumn{1}{c|}{Overall} &\multicolumn{5}{c|}{Huggingface Open LLM Leaderboard} & \multicolumn{3}{c}{InstructEval} \\
 & Average & HellaSwag & ARC & TruthfulQA & MMLU & RTE & BBH & DROP & HumanEval \\
\hline
Diversity-Only & 42.48 & 79.26 & 46.67 & 35.49 & 45.04 & 66.43 & 33.12 & 21.77 & 12.2 \\
Non-Iterative & 40.48 & 79.2 & 47.35 & 35.86 & 44.87 & 57.76 & 33.4 & 11.36 & 14.02 \\
Random Selection & 41.62 & 79.32 & \textbf{48.89} & 35.68 & 42.88 & 56.68 & 33.75 & 24.15 & 11.59 \\
Score-Only & 43.77 & 79.35 & 47.87 & 37.96 & 39.56 & 72.56 & 33.33 & 26.73 & 12.8 \\
Ours & \textbf{46.15} & \textbf{79.9} & 47.44 & \textbf{38.58} & \textbf{45.53} & \textbf{78.7} & \textbf{33.78} & \textbf{28.81} & \textbf{16.46} \\
\hline
\end{tabular}
}
\caption{Comparison of performance across different ablation models using 20\% of the data on the Huggingface Open LLM Leaderboard and InstructEval Leaderboard.}
\label{table:comparison_basic_ablation}
\end{table*}

\begin{table}[ht]
\centering
\resizebox{0.5\textwidth}{!}{
\begin{tabular}{l|ccc|cc}
\hline
 & \multicolumn{3}{c|}{MT Bench} & \multicolumn{2}{c}{AlpacaEval 2.0} \\
 & Overall & turn1 & turn2 & \makecell{length controlled \\ win rate} & win rate \\
\hline
Diversity-Only & 4.884 & 5.606 & 4.163 & 3.68 & 1.71 \\
Non-Iterative & 5.066 & 5.894 & 4.238 & 4.02 & 1.83 \\
Random Selection & 4.728 & 5.738 & 3.719 & 3.78 & 1.58 \\
Score-Only & 4.988 & 5.919 & 4.056 & 4.6 & 2.4 \\
Ours & \textbf{5.228} & \textbf{6.194} & \textbf{4.263} & \textbf{4.92} & \textbf{2.65} \\
\hline
\end{tabular}
}
\caption{Comparison of performance across different ablation models using 20\% of the data on MT Bench and AlpacaEval 2.0.}
\label{table:comparison_instrucion_ablation}
\end{table}

\subsection{Ablations on the Base Model}

The choice of base model is crucial to the performance of fine-tuned models. While our primary experiments use LLaMA2-7B, we also evaluate our approach using more powerful models, LLaMA2-13B, and LLaMA3.1-8B, to assess its robustness. For each model, we apply our data selection method on 20\% of the data and compare the results with full-data fine-tuning.

As shown in Appendix~\ref{sec:llama2-13b-llama3.1-8b}, both models improve over LLaMA2-7B, highlighting the impact of using a stronger base model. The 20\% data fine-tuned models outperform their full-data counterparts, though the performance gap narrows with these models, suggesting that stronger base models are less sensitive to fine-tuning data volume with our method.
Additionally, LLaMA3.1-8B achieves the best overall performance, underscoring the significance of base model strength in fine-tuning.

\section{Related Work}

\subsection{Instruction Fine-Tuning}


Instruction fine-tuning has proven to be an effective method for improving large language models' (LLMs) ability to understand and follow natural language instructions. This process involves fine-tuning pre-trained models on datasets \(\mathcal{D} = \{(x_i, y_i)\}_{i=1}^{N}\), where \(x_i\) represents an instruction and \(y_i\) the corresponding response. Early work, such as that with GPT-3 \cite{BrownMRSKDNSSAA20}, highlighted the broad task improvement achieved through this approach. Recent models, including LLaMA \cite{touvron2023llama} and Alpaca \cite{taori2023alpaca}, have refined this process, emphasizing the selection of high-quality instruction pairs to improve generalization and aligning model outputs more closely with human expectations.

\subsection{Instruction-Tuning Data Selection}

Several methods have been developed to efficiently select high-quality instruction-tuning data. \citet{ChenLYWGYTS0HJ24} utilized a ChatGPT-based evaluator to filter responses based on accuracy and relevance. \citet{LiZLCC0W0024} introduced Instruction-Following Difficulty (IFD) scores, which measure the loss difference between an instruction-response pair and its direct response, thereby identifying more challenging data. \citet{cao2023instruction} leveraged inference loss and natural language indicators to estimate instruction quality, while \citet{LiHXYYZSCLLHL24} proposed a one-shot improvement metric that classifies high-quality data based on its ability to significantly enhance performance in one-shot settings. \citet{chen2023maybe} employed a coreset-based approach, selecting high-quality data by identifying core samples post-clustering.

In contrast, our approach directly evaluates whether the base model can effectively handle each instruction using GPT-4's judgment and trains a smaller classifier to mimic GPT-4's evaluations. While some works \cite{MekalaNS24,LiZHLZWCZ24,LiCCHGZ24} have also explored the use of smaller models for efficient instruction data selection, our method primarily focuses on identifying instruction data that the base LLM struggles to handle, distinguishing it from prior approaches.

\section{Conclusion}


We introduce an iterative training policy framework for efficiently selecting high-quality instruction-tuning data, requiring no human involvement and minimal use of GPT-4. Our approach demonstrates that fine-tuning a model with approximately 20\% of the chosen data from the source set consistently outperforms models fine-tuned on the full dataset. In experiments with Alpaca and WizardLM, our method demonstrates strong performance with reduced data volumes (5\% for Alpaca, 10\% with WizardLM) compared to the original full-data model. Ablation studies across different base LLMs and the exclusion of key components demonstrate the robustness and effectiveness of our policy.



\section*{Limitations}

There are two primary limitations to consider in our work. First, in constructing the source set \(\mathcal{S}\), we randomly sample 15,000 instruction data from each source for diversity without thoroughly evaluating data quality within each source. Future research could consider curating a more optimized and high-quality source set for fine-tuning. Second, in the k-means clustering step, we do not explore all possible configurations for the number of clusters and the number of samples selected per cluster. Future studies could investigate the impact of different k-means parameters on the diversity and effectiveness of the selected instruction data.





\bibliography{custom}

\appendix

\clearpage

\section{The Algorithm Workflow}
\label{sec:appendix_algoritm_workflow}

\subsection{Traning Stage Workflow}
Detailed algorithm workflow of the training stage is shown in Algorithm \ref{training flow}.
\label{sec:appendix_training_workflow}

\begin{algorithm}
\small
\caption{ Training Stage Workflow}
\label{training flow}
\KwIn{Source set $\mathcal{S} = (\mathcal{X}, \mathcal{Y})$, fixed batch size \( B \)}
\KwOut{Trained BERT classifier model $f$}
\For{iteration $i = 0$ \textbf{to} $n$}{
    Select a diverse subset $\mathcal{V}_i$ using K-means clustering from source set $\mathcal{S}_i$\;
    \eIf{i=0}{
        $\mathcal{D}_0 \leftarrow$Randomly select $B$ samples from $\mathcal{V}_i$ without scoring\;
    }{
        Calculate score $Q_{i}$ via ~\autoref{eq:4}\;
        $\mathcal{D}_i \leftarrow$ Select top $B$ instruction samples from $\mathcal{V}_i$\;
    }
    Use base LLM to generate answers $\hat{\mathcal{Y}}_i$ for instructions $\mathcal{X}_i$ \;
    $(\mathcal{D}_{i}^{\text{hard}}, \mathcal{D}_{i}^{\text{easy}}) \leftarrow$ Evaluate response $(\hat{\mathcal{Y}}_i, \mathcal{Y}_i)$ with GPT-4 via ~\autoref{eq:5} \;
    Iterative train BERT model $f$ using dataset $(\mathcal{D}_{i}^{\text{hard}}, \mathcal{D}_{i}^{\text{easy}})$\;
    \If{validation accuracy $>$ 95\%}{
        \textbf{break}\;
    }
    $\mathcal{D}^{\text{H}}_{i+1} \gets \mathcal{D}_{i}^{\text{H}} \cup \mathcal{D}_{i}^{\text{hard}}$\;
    $\mathcal{S}_{i+1} \gets \mathcal{S}_i \setminus \mathcal{D}_{i}$\;
}
\end{algorithm}

\subsection{Inference Stage Workflow}
Detailed algorithm workflow of the inference stage is shown in Algorithm \ref{Inference flow}.
\label{sec:appendix_inference_workflow}

\begin{algorithm}
\small
\caption{ Inference Stage Workflow}
\KwIn{Remaining Source set $\mathcal{S}_{i+1} = (\mathcal{X}, \mathcal{Y})$, trained classifier ${f}$, hard dataset $\mathcal{D}^{\text{H}}$, selection rate $\alpha$}
\KwOut{Selected fine-tuning data $\mathcal{D}_{\text{final}}$}
$N_{\text{sel}} \leftarrow |\mathcal{S}_{i+1}| \times \alpha$;  \tcp{Calculate data amount} 
$\mathcal{V} \gets$ Use k-means to obtain a diverse subset ; \tcp{$|\mathcal{V}| = 3\times N_{sel}$} 
Calculate score $Q$ via ~\autoref{eq:4} \;
$\mathcal{D} \gets \{\, \mathcal{V}_{(1)}, \mathcal{V}_{(2)}, \ldots, \mathcal{V}_{(N_{sel})}\}$ where $Q(\mathcal{V}_{(1)}) \geq Q(\mathcal{V}_{(2)}) \geq \cdots \geq Q(\mathcal{V}_{(N_{sel})}) $\;
$\mathcal{D}_{\text{final}} \gets \mathcal{D} \cup \mathcal{D}^{\text{H}}$;
\label{Inference flow}
\end{algorithm}


\section{Iterative Training Results of the Classifier}
\label{sec:appendix_classifier_training}

To assess the classifier's performance during iterative training, we track two key metrics: the number of "easy/hard" instructions and the validation accuracy. The "easy/hard" instructions indicate how many instructions GPT-4 classified as "hard" or "easy" from the fixed number of selected instructions \(\mathcal{D}\) during each iteration. Validation accuracy reflects the classifier's accuracy on the validation set at each iteration.

As shown in Table~\ref{tab:iteration_performance}, the classifier is trained iteratively, with each iteration demonstrating an increase in both the number of "hard" instructions identified and the validation accuracy. This indicates an improvement in the classifier's ability to identify challenging instructions over time, enhancing overall model performance.


\begin{table}[ht]
    \centering
    \caption{Classifier Performance Across Iterations}
    \scriptsize
    \resizebox{0.5\textwidth}{!}{ 
    \begin{tabular}{@{}cccc@{}}
    \toprule
    \textbf{Iteration} & \textbf{Hard Instructions} & \textbf{Easy Instructions} & \textbf{Validation Accuracy (\%)} \\ \midrule
    0                  & 338                        & 62                         & 81.2                              \\
    1                  & 368                        & 32                         & 87.87                             \\
    2                  & 377                        & 23                         & 91.67                             \\
    3                  & 381                        & 19                         & 96.87                             \\ \bottomrule
    \end{tabular}}
    \label{tab:iteration_performance}
\end{table}

In the initial iteration, GPT-4 identifies 338 instructions as "hard", with the classifier achieving a validation accuracy of 81.2\%. As the iterations progress, both the number of "hard" instructions and validation accuracy steadily increase. By the final iteration, GPT-4 classifies 381 instructions as "hard", and the validation accuracy reaches 96.87\%, demonstrating the model's growing proficiency in aligning with GPT-4's judgments.

\section{Analysis of the Weighting Factor \( \alpha \)}
\label{sec:alpha_study}

We evaluate different values of \( \alpha \), ranging from 0.6 to 0.9, to assess their impact on the model's ability to identify challenging instructions.

\begin{figure}
    \centering
    \includegraphics[width=1.0\linewidth]{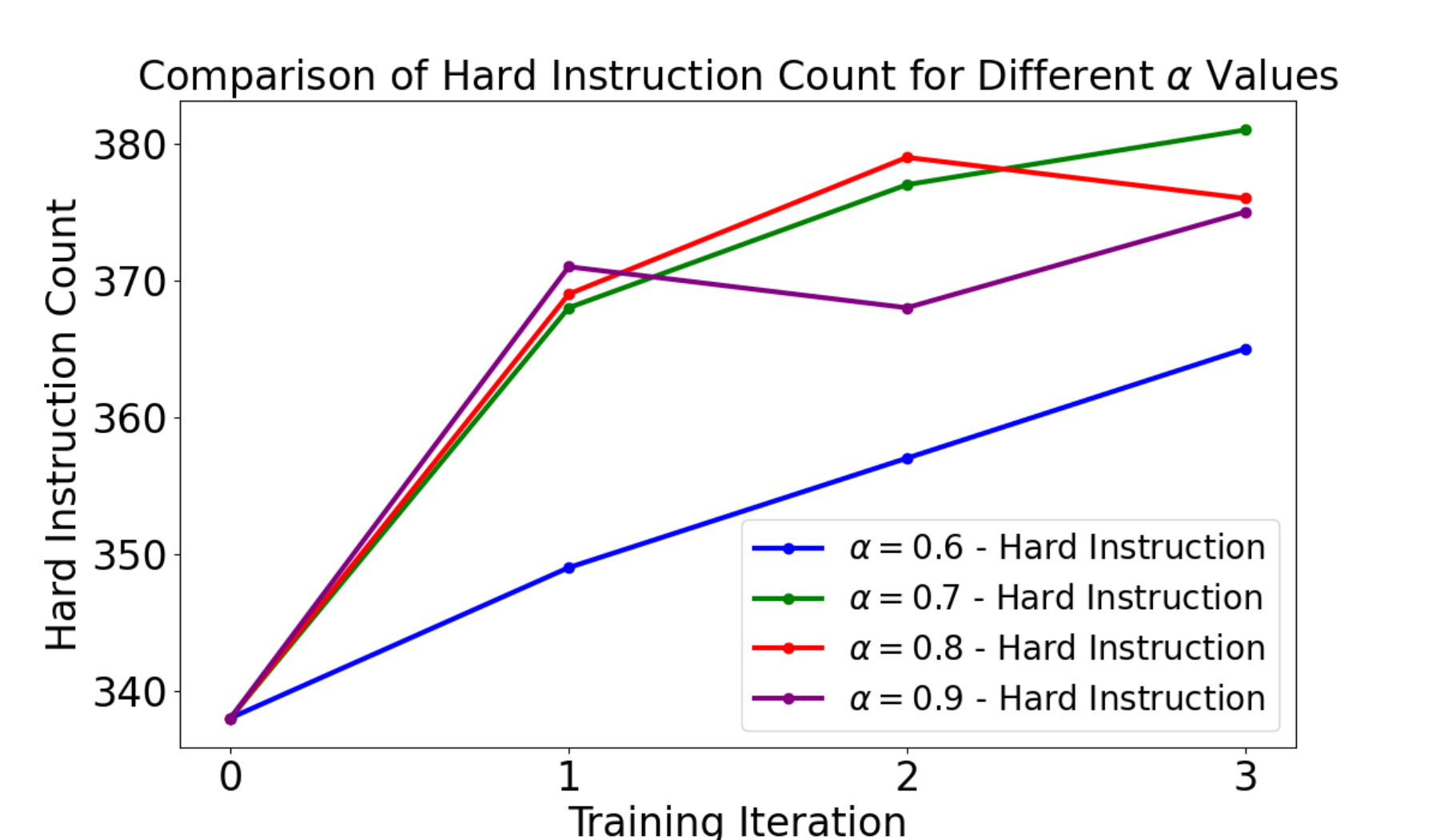}
    \caption{Comparison of the number of "hard" instructions identified across iterations for different \( \alpha \). Results shown up to iteration 3.}
    \label{fig:alpha_0.7}
\end{figure}

Figure~\ref{fig:alpha_0.7} compares the number of "hard" instructions identified by GPT-4 across iterations for each value of \( \alpha \). In the initial iteration (iteration 0), 400 instructions are randomly selected without applying the scoring mechanism, resulting in all curves starting from the same point.

The results show that while all values of \( \alpha \) lead to an increase in "hard" instructions in the early iterations, higher values such as \( \alpha = 0.8 \) and \( \alpha = 0.9 \) cause a performance decline in later iterations. In contrast, \( \alpha = 0.6 \) and \( \alpha = 0.7 \) display a consistent, monotonic increase in the number of "hard" instructions, with \( \alpha = 0.7 \) yielding the best overall performance.

Based on these findings, we select \( \alpha = 0.7 \) as the optimal weighting factor, providing a balanced contribution from both the classifier and similarity, leading to more effective data selection.

\section{Prompt for Evaluation}
\label{sec:prompt_evaluation}

In Table \ref{table:prompt_gpt4}, we provide the detailed prompt we used for evaluating the performance of two responses for the same instruction.

\begin{center}
\begin{minipage}{0.5\textwidth} 
\setlength{\tabcolsep}{6pt} 
\begin{tabular}{p{0.95\textwidth}}
\hline


\textbf{System Prompt} \\
You are a helpful and precise assistant for checking the quality of the answer. \\
\\
\textbf{User Prompt} \\
\texttt{[Question]} \\
\textit{Question} \\
\texttt{[The Start of Assistant 1's Answer]} \\
\textit{Answer 1} \\
\texttt{[The End of Assistant 1's Answer]} \\
\texttt{[The Start of Assistant 2's Answer]} \\
\textit{Answer 2} \\
\texttt{[The End of Assistant 2's Answer]} \\
\\
We would like to request your feedback on the performance of two AI assistants in response to the user question displayed above. Please rate the helpfulness, relevance, accuracy, and level of detail of their responses. Each assistant receives an overall score on a scale of 1 to 10, where a higher score indicates better overall performance. \\
Please first output a single line containing only two values indicating the scores for Assistant 1 and Assistant 2, respectively. The two scores are separated by a space. In the subsequent line, please provide a comprehensive explanation of your evaluation, avoiding any potential bias and ensuring that the order in which the responses were presented does not affect your judgment. \\

\hline
\end{tabular}
\end{minipage}
\end{center}

\begin{table}[h!]
\centering
\caption{The prompt we use to request GPT-4 to evaluate the responses.}
\label{table:prompt_gpt4}
\end{table}

\section{Fine-tuning Settings}
\label{sec:appendix_implementation_details}

Fine-tuning is performed using the
Alpaca codebase\footnote{\url{https://github.com/tatsu-lab/stanford_alpaca}} with DeepSpeed ZeRO-2 \cite{rasley2020deepspeed} for optimization. The learning rate is set to \(2 \times 10^{-5}\) with a warmup ratio of 0.03, following a cosine decay schedule. The maximum token length is 1024, and training is conducted using bf16 precision \cite{burgess2019bfloat16}. The model is fine-tuned for 3 epochs with a batch size of 128.

\section{Detailed Comparisons on Test Set}
\label{sec:appendix_extra_5_15_60}

Comparisons of Win/Tie/Lose for models fine-tuned on 5\%, 15\%, and 60\% of the data with full-data fine-tuned model are shown below in Figure \ref{figure:three_datasets}. Results for 10\% and 20\% data fine-tuning are provided in the main paper.

\begin{figure*}[t]
    \centering
    \subfloat[5\% fine-tuning data]{%
        \includegraphics[width=0.32\linewidth]{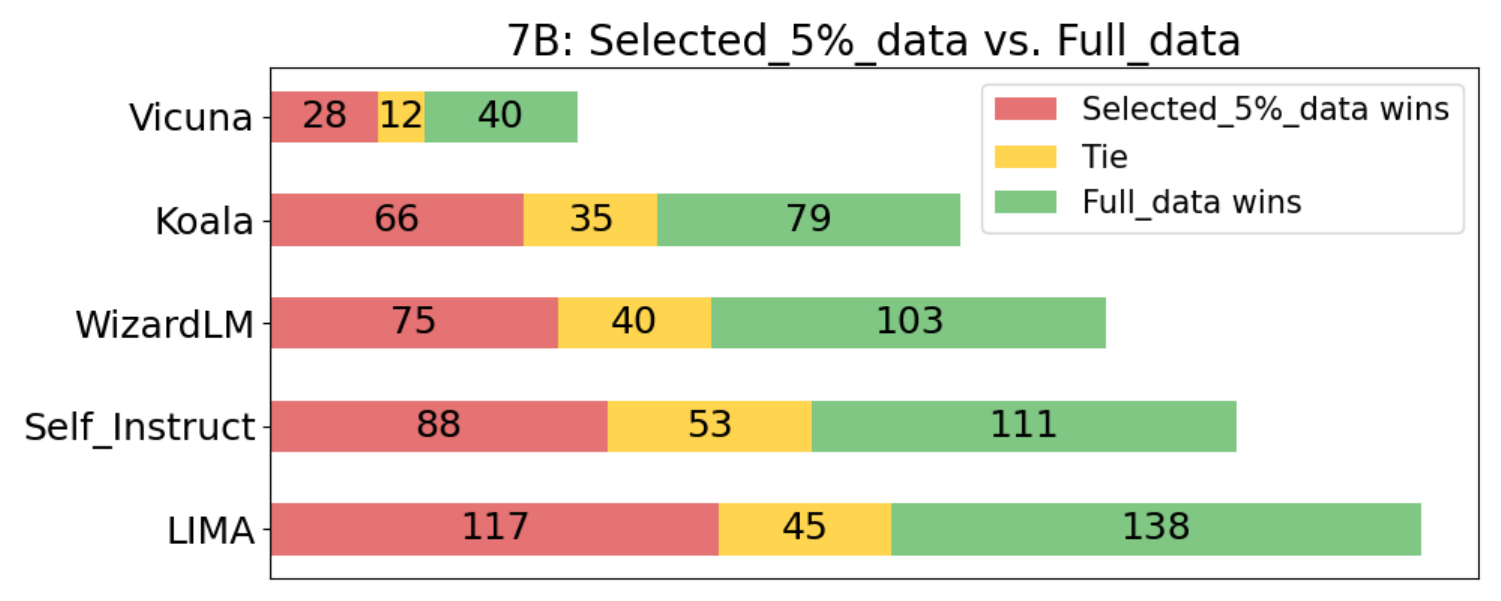}
    }\hfill
    \subfloat[15\% fine-tuning data]{%
        \includegraphics[width=0.32\linewidth]{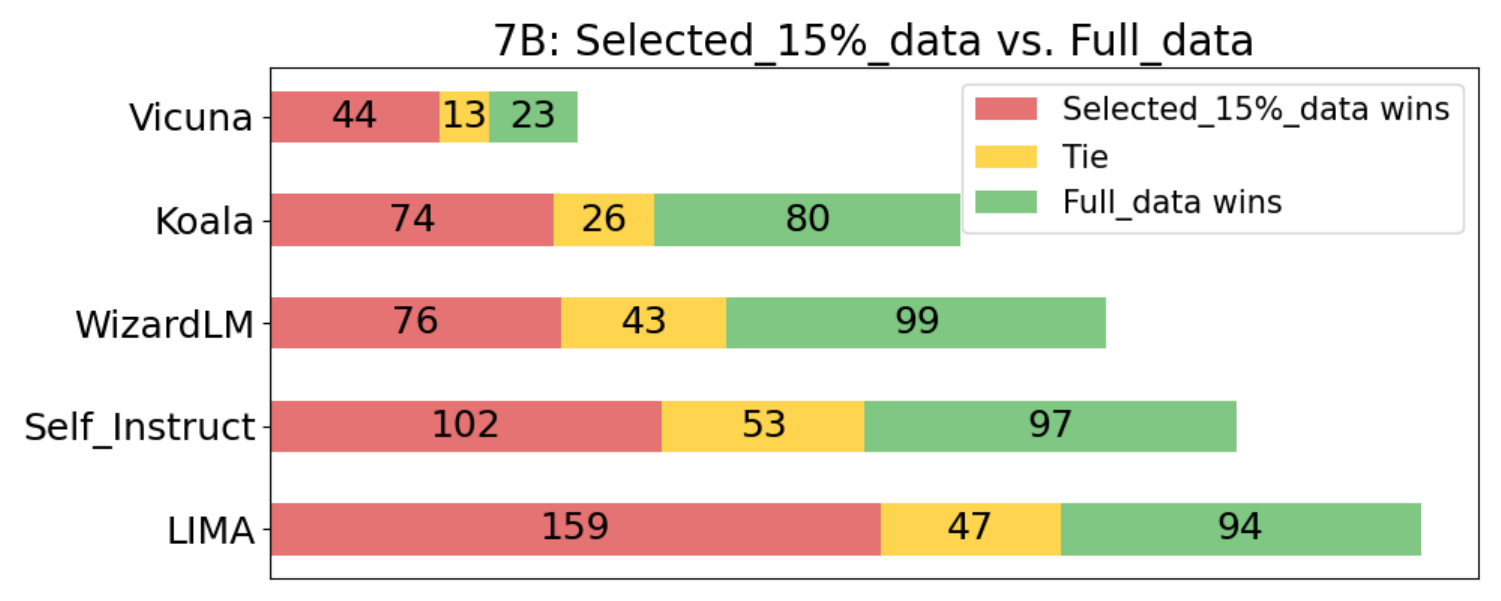}
    }\hfill
    \subfloat[60\% fine-tuning data]{%
        \includegraphics[width=0.32\linewidth]{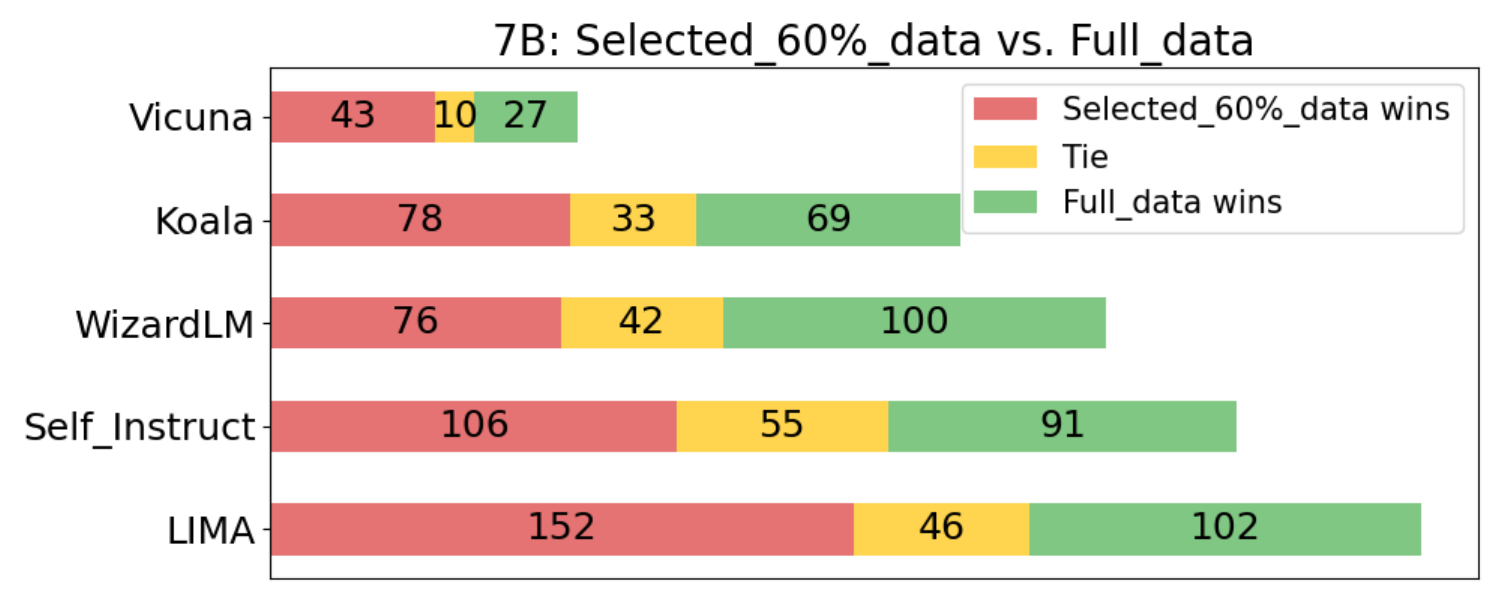}
    }
    \caption{Comparisons of Win/Tie/Lose for models fine-tuned on 5\%, 15\%, and 60\% of the data with the full-data fine-tuned model.}
    \label{figure:three_datasets}
\end{figure*}

\section{Implementation Details of Alpaca and WizardLM}
\label{sec:appendix_alpaca_wizardlm_implementation_details}

The Alpaca dataset consists of 52,000 instruction-response pairs, while the WizardLM contains 70,000 pairs. Following the setup in the main paper, where 5\% of Alpaca data and 10\% of WizardLM data are selected for fine-tuning, we choose 2,600 instruction pairs from Alpaca and 7,000 pairs from WizardLM for the fine-tuning process.

For the diverse instruction subset \( \mathcal{V} \), we set the size to 10 times the final selected Alpaca data and 5 times the final selected WizardLM data. K-means clustering is applied with 100 clusters to ensure diversity in the selected subset.

In contrast to the inference mode used in the main experiments, the cumulative "hard" instructions are not treated as default chosen high-quality data. Instead, they are utilized solely for calculating the similarity score. After constructing the diverse subset \( \mathcal{V} \), we directly apply the inference mode of our policy to select the top-scoring instructions for fine-tuning (2,600 for Alpaca and 7,000 for WizardLM).

All other experimental settings follow the same as outlined in \cite{LiZLCC0W0024}.

\section{Test Set Comparison: Ablation Models vs. Our Model}

\label{sec:appendix_testset_ablation}

\begin{figure*}[t]
    \centering
    \subfloat[Ours vs. Diversity-only]{%
        \includegraphics[width=0.48\linewidth]{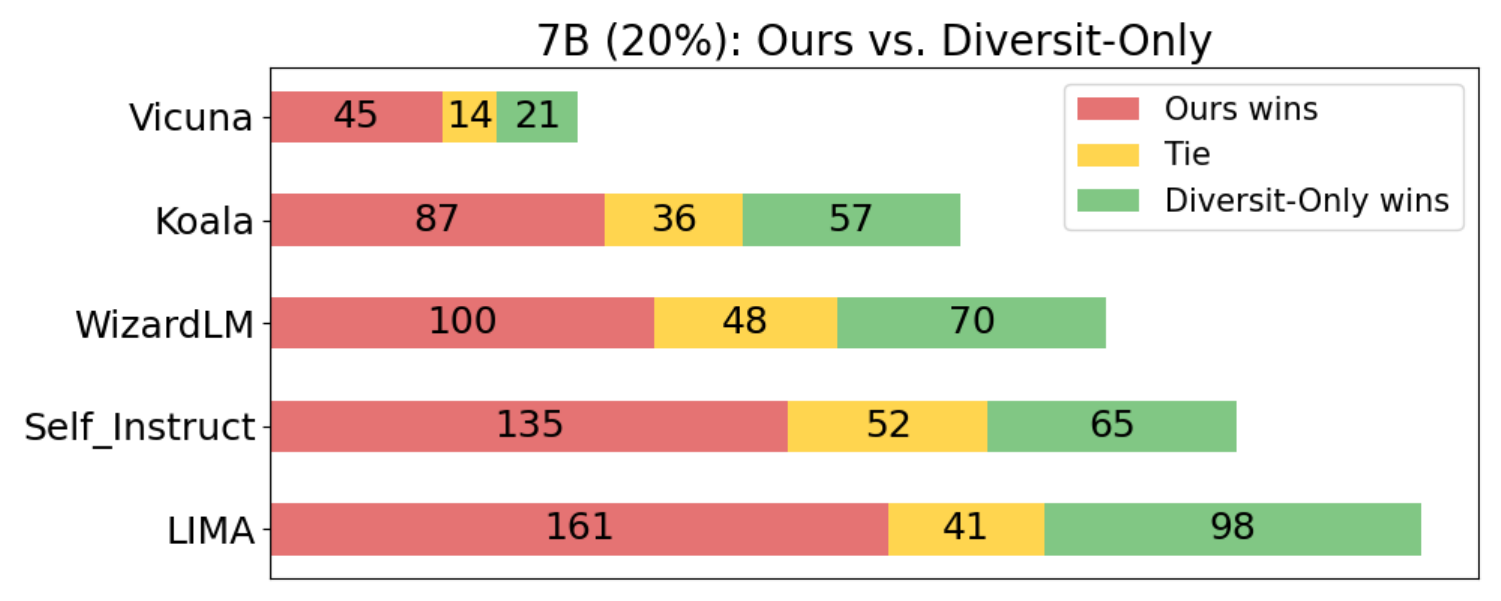}
    }\hfill
    \subfloat[Ours vs. Non-Iterative]{%
        \includegraphics[width=0.48\linewidth]{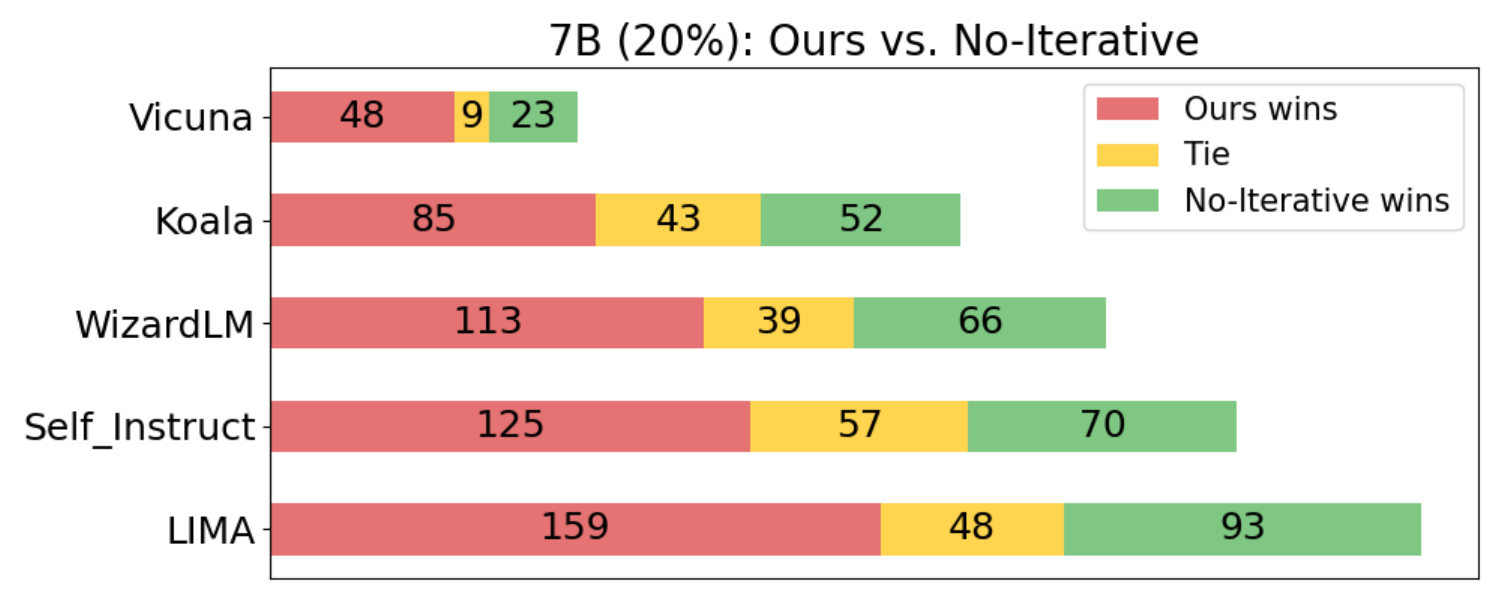}
    }\hfill
    \subfloat[Ours vs. Random-selection]{%
        \includegraphics[width=0.48\linewidth]{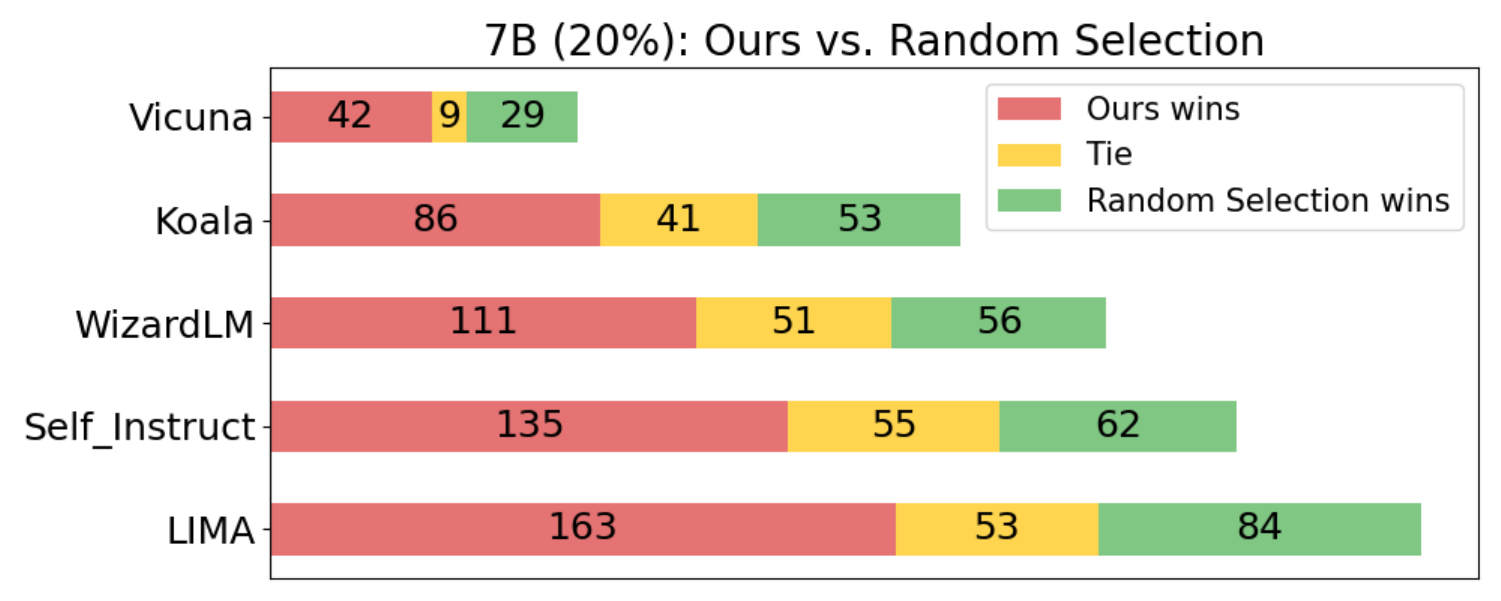}
    }\hfill
    \subfloat[Ours vs. Score-only]{%
        \includegraphics[width=0.48\linewidth]{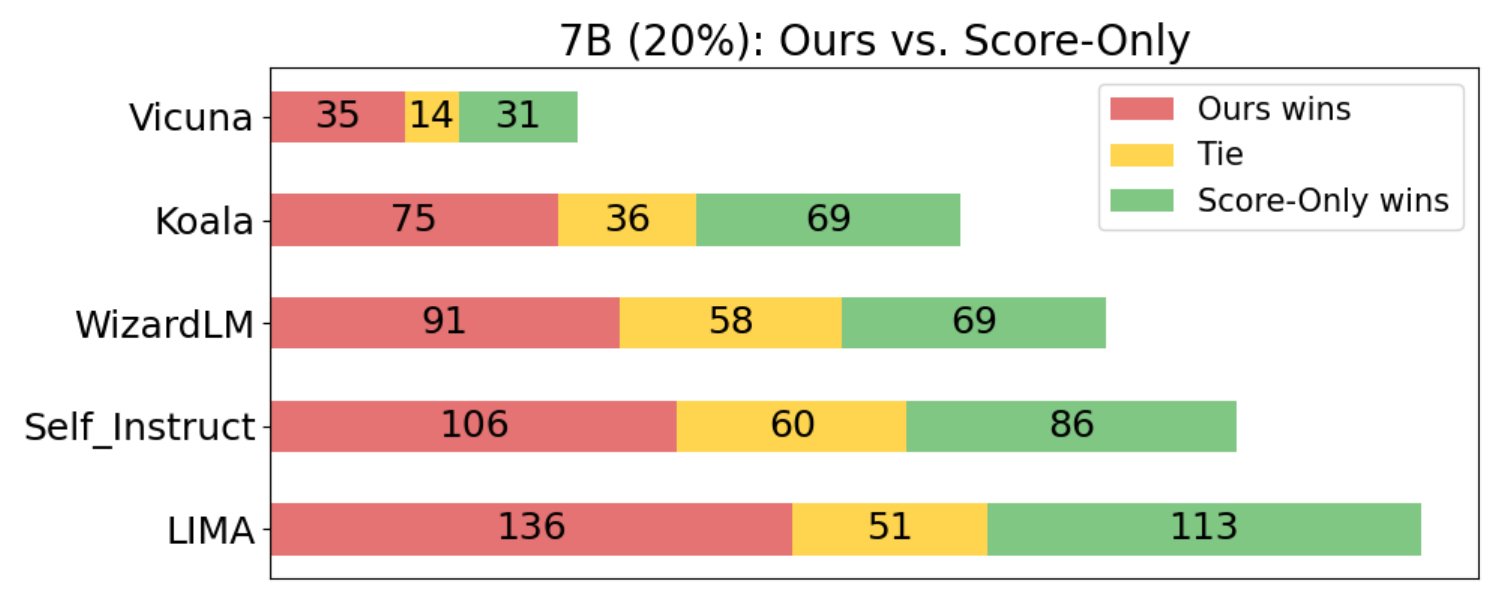}
    }
    \caption{Comparison of Win/Tie/Lose between 20\% data fine-tuned model of ours and different ablation methods.}
    \label{figure:20_ablation}
\end{figure*}

Figure \ref{figure:20_ablation} presents the Win/Tie/Lose comparison on different test sets between our 20\% fine-tuned model and the various ablation methods. The results clearly demonstrate that our model consistently outperforms all ablation models across all test sets, highlighting the effectiveness of our approach. Notably, the performance gap between our model and the \texttt{score-only} model is the smallest among the four ablation methods, underscoring the importance of the scoring mechanism. In contrast, the \texttt{random-selection} model shows the largest performance gap compared to our method, further validating the overall success of our data selection framework in identifying high-quality data.


\section{Detailed Evaluation Results on LLAMA2-13B and LLAMA3.1-8B}
\label{sec:llama2-13b-llama3.1-8b}
\begin{figure*}[t]
    \centering
    \subfloat[LLaMA2-13B]{%
        \includegraphics[width=0.48\linewidth]{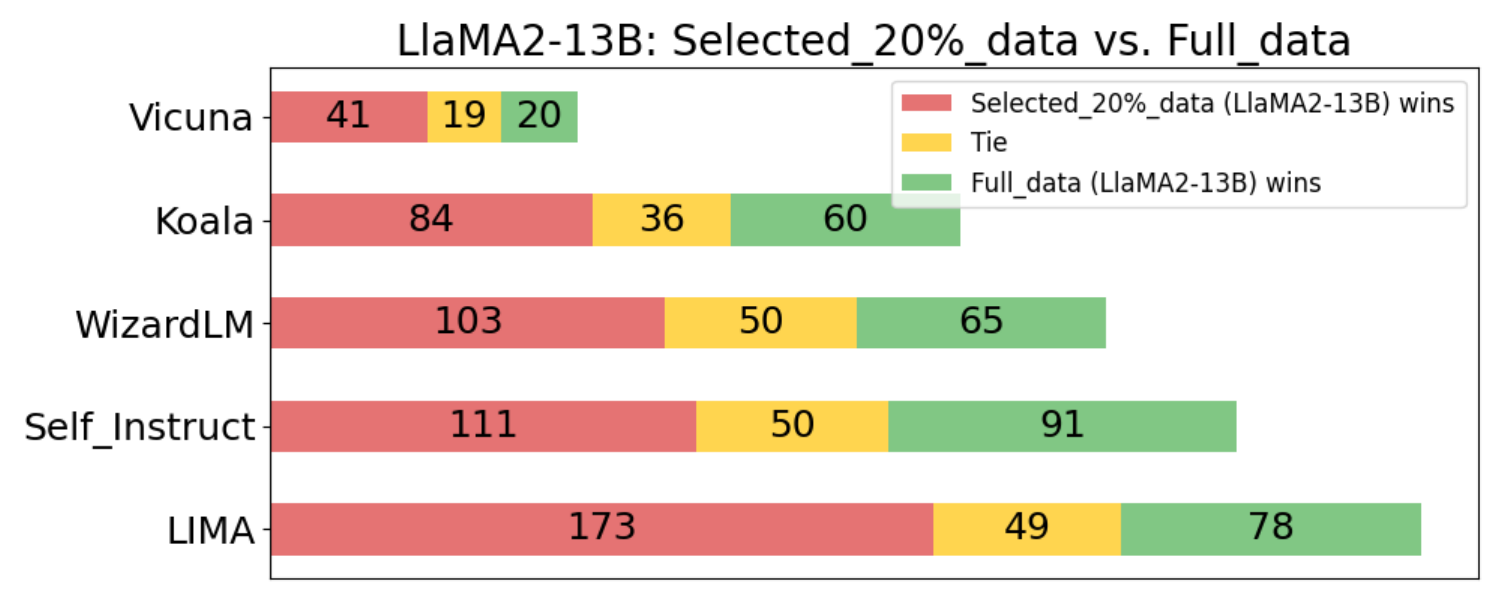}
    }\hfill
    \subfloat[LLaMA3.1-8B]{%
        \includegraphics[width=0.48\linewidth]{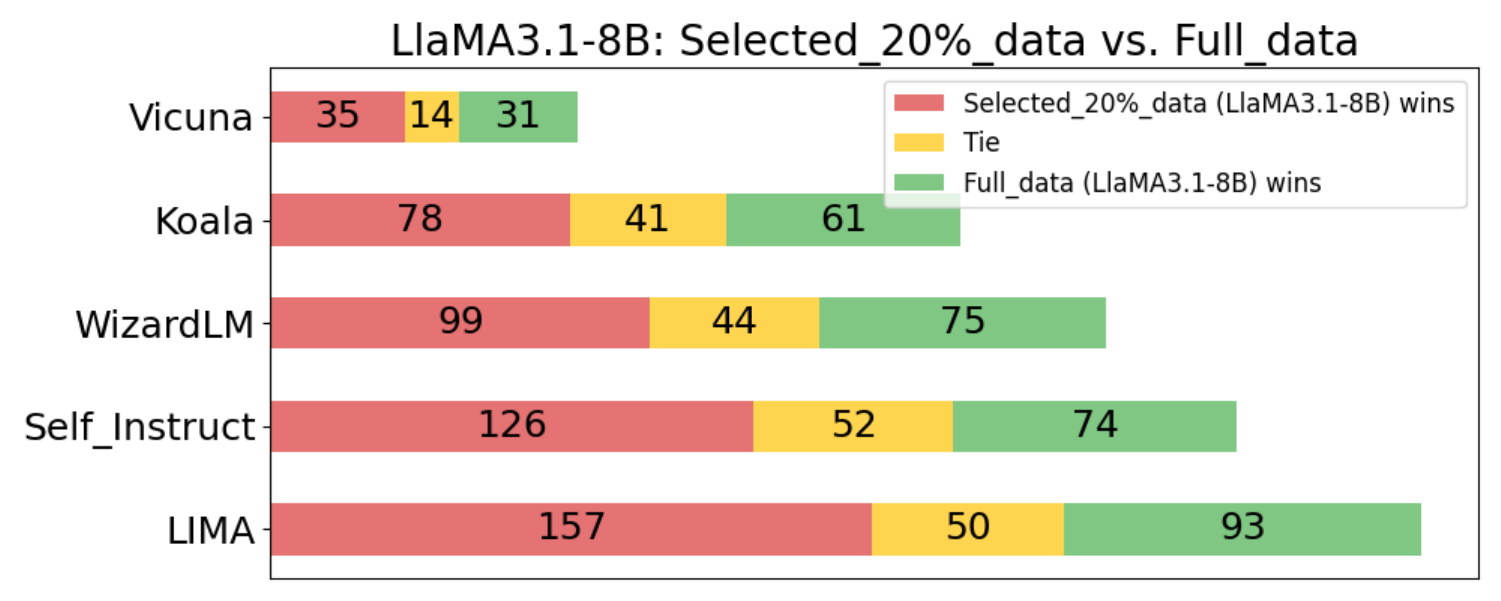}
    }
    \caption{Comparison of Win/Tie/Lose between our 20\% data fine-tuned model and full-data fine-tuned model with different base models: LLaMA2-13B (left) and LLaMA3.1-8B (right).}
    \label{figure:LLaMA2-13b-3.1-8b}
\end{figure*}

Benchmark results and test set comparisons of the selected 20\% data fine-tuned model and full-data fine-tuned model using base model LLaMA2-13B and LLaMA3.1-8B are shown in Table \ref{table:comparison_basic_LLaMA2-13b-3.1-8b}, Table \ref{table:comparison_instruction_LLaMA2-13b-3.1-8b} and Figure \ref{figure:LLaMA2-13b-3.1-8b}.

\begin{table*}[ht]
\centering
\resizebox{\textwidth}{!}{
\begin{tabular}{l|c|ccccc|ccc}
\hline
 & \multicolumn{1}{c|}{Overall} &\multicolumn{5}{c|}{Huggingface Open LLM Leaderboard} & \multicolumn{3}{c}{InstructEval} \\
 & Average & HellaSwag & ARC & TruthfulQA & MMLU & RTE & BBH & DROP & HumanEval \\
\hline
Selected\_20\%\_data (LLaMA2-13B) & \textbf{49.24} & \textbf{82.57} & 50.6 & \textbf{35.98} & \textbf{52.63} & 77.98 & \textbf{38.69} & 39.62 & \textbf{15.85} \\
Full\_data (LLaMA2-13B) & 49.21 & 81.63 & \textbf{51.71}  & 35.79 & 52.39 & \textbf{78.34} & 38.46 & \textbf{40.14} & 15.24 \\
\hline
Selected\_20\%\_data (LLaMA3.1-8B) & \textbf{53.00} & \textbf{81.73} & \textbf{53.5} & 38.81 & \textbf{57.95} & 74.01 & \textbf{42.09} & \textbf{44.82} & \textbf{31.1} \\
Full\_data (LLaMA3.1-8B) & 52.14 & 80.22 & 51.54  & \textbf{40.86} & 54.76 & \textbf{79.42} & 40.64 & 42.88 & 26.83 \\
\hline
\end{tabular}
}
\caption{The comparison of the performance of LLaMA2-13B and LLaMA3.1-8B on Huggingface Open LLM Leaderboard and InstructEval Leaderboard.}
\label{table:comparison_basic_LLaMA2-13b-3.1-8b}
\end{table*}

\begin{table*}[ht]
\centering
\begin{tabular}{l|ccc|cc}
\hline
 & \multicolumn{3}{c|}{MT Bench} & \multicolumn{2}{c}{AlpacaEval 2.0} \\
 & Overall & turn1 & turn2 & length controlled win rate & win rate \\
\hline
Selected\_20\%\_data (LLaMA2-13B) & \textbf{5.681} & \textbf{6.5} & 4.863 & \textbf{5.15} & \textbf{2.47} \\
Full\_data (LLaMA2-13B) & 5.563 & 6.213 & \textbf{4.913} & 4.65 & 2.2 \\
\hline
Selected\_20\%\_data (LLaMA3.1-8B) & \textbf{5.8} & \textbf{6.763} & 4.838 & \textbf{6.6} & \textbf{3.24} \\
Full\_data (LLaMA3.1-8B) & 5.519 & 6.131 & \textbf{4.906} & 4.8 & 2.09 \\
\hline
\end{tabular}
\caption{The comparison of performance of LLaMA2-13b and LLaMA3.1-8B on MT Bench and AlpacaEval 2.0.}
\label{table:comparison_instruction_LLaMA2-13b-3.1-8b}
\end{table*}

\end{document}